\documentclass{article}

\usepackage[final]{acl}
\usepackage{natbib}
\usepackage{fontawesome5}  

\usepackage[utf8]{inputenc}
\usepackage[T1]{fontenc}

\usepackage{amsmath, amssymb, amsfonts}

\usepackage{parskip}
\usepackage{enumitem}
\usepackage{float}
\usepackage{array}
\usepackage{nicefrac}

\usepackage{booktabs}
\usepackage{tabularx}
\usepackage{makecell}
\usepackage{multirow}
\usepackage{colortbl}
\usepackage{graphicx}
\graphicspath{{./images/}}

\usepackage{xcolor}
\definecolor{navyblue}{RGB}{31,56,100}
\definecolor{steelblue}{RGB}{46,84,150}
\definecolor{lightgray}{RGB}{245,245,245}
\definecolor{codelabel}{RGB}{80,80,80}
\definecolor{rowalt}{RGB}{240,245,253}
\definecolor{bestrow}{RGB}{220,235,220}
\definecolor{darkgreen}{RGB}{34,110,34}
\definecolor{darkred}{RGB}{160,30,30}

\usepackage{hyperref}
\usepackage{url}
\hypersetup{colorlinks=true, linkcolor=steelblue, citecolor=steelblue, urlcolor=steelblue}

\usepackage{caption}
\usepackage{listings}
\usepackage{mdframed}
\newmdenv[
  backgroundcolor=rowalt,
  linecolor=steelblue,
  linewidth=1pt,
  leftmargin=0pt,
  rightmargin=0pt,
  innerleftmargin=10pt,
  innerrightmargin=10pt,
  innertopmargin=8pt,
  innerbottommargin=8pt,
  skipabove=6pt,
  skipbelow=6pt
]{pipelinebox}

\newcommand{\up}[1]{{\color{darkgreen}$\uparrow$\,#1}}

\newcommand{\sys}[1]{\textsc{#1}}
\newcommand{\tblbest}[1]{\textbf{#1}}

\title{CaVe-VLM-CoT: An Interpretable Vision-Language Model Framework}
\author{
\normalsize
Sneha Rao \quad Shaina Raza \quad Dhanesh Ramachandram\\
\small
Vector Institute, Toronto, ON M5G 0C6, Canada\\
\texttt{\{sneha.rao, shaina.raza, dhanesh.ramachandram\}@vectorinstitute.ai}
}

\begin{document}
\maketitle

\begin{abstract}
Vision-Language Models (VLMs) remain prone to hallucinations, producing fluent but visually unfaithful outputs. Existing chain-of-thought and retrieval-augmented methods only partially address this, as they neither enforce step-level citation grounding nor route verification failures back to retrieval for correction. We present CaVe-VLM-CoT, a modular reflection-based agentic-RAG framework that enforces evidence-grounded reasoning  through a five-stage closed-loop pipeline: Extractor, Retriever, Solver, Citation Injector, and Verifier, in which detected ungrounded claims trigger structured feedback to the Extractor for targeted re-retrieval. Since no existing framework jointly measures retrieval quality, step-wise citation faithfulness, and cross-modal grounding, we propose a suite of 23 component-wise metrics across all stages, anchored by CaVeScore, a composite metric weighting accuracy, citation precision and recall, attribution, and evidence grounding. Without any architectural or prompt modifications, CaVe-VLM-CoT achieves 87.1\% accuracy and 56.6\% CaVeScore on ScienceQA , and 55.2\% accuracy and 35.7\% CaVeScore on MMMU (30 subjects). 
We release the code for reproducibility \quad \faGithub~\href{https://anonymous.4open.science/r/cave-vlm-cot-062A/README.md}{Project code} 
\end{abstract}

\section{Introduction}
VLMs take an image and a question and produce an answer through some form of reasoning. Systems have advanced considerably from early VQA models~\citep{antol2015vqa} to instruction-tuned multimodal models such as LLaVA~\citep{liu2023llava}, GPT-4V~\citep{openai2023gpt4v}, and InstructBLIP~\citep{dai2023instructblip}. Despite so much progress, these models frequently generate fluent but unfaithful responses, hallucinating objects, evidence, and reasoning steps unsupported by the visual or textual context~\citep{li2023evaluating, bai2024hallucination}. Reported hallucination rates exceed 40\% on knowledge-intensive queries~\citep{wu2025generate}, a serious concern in domains such as medicine, finance, and education, where errors are costly.

Two research directions partially address this problem. Chain-of-thought (CoT) prompting~\citep{wei2022chain} exposes intermediate reasoning steps, with multimodal extensions such as LLaVA-CoT~\citep{xu2025llava} adding structured output formats or post-hoc visual grounding. Retrieval-augmented generation (RAG)~\citep{lewis2020retrieval} grounds outputs in retrieved evidence, with multimodal variants such as RIV-CoT~\citep{corbiere2025retrieval} and MI-RAG~\citep{choi2025multimodal} retrieving evidence during reasoning. However, these approaches leave three important gaps unaddressed: (1) the absence of step-level citation mechanisms that require each reasoning step to be grounded in specific retrieved evidence; (2) the lack of an explicit verification stage to determine whether retrieved context has been faithfully incorporated into the reasoning process; and (3) the absence of a self-correcting feedback loop capable of tracing ungrounded claims back to the \textsc{Extractor} for corrective re-querying and re-retrieval.

In this work, we present \textbf{CaVe-VLM-CoT} (Cite-and-Verify Vision-Language Model with Chain-of-Thought), a modular reflection-based agentic-RAG framework that addresses all three gaps simultaneously. Rather than treating retrieval, reasoning, and verification as separate concerns, it integrates them into a  five-stage closed-loop pipeline comprising an \textsc{Extractor}, \textsc{Retriever}, \textsc{Solver}, \textsc{Citation Injector} and \textsc{Verifier}, looping until reasoning is verifiably grounded or a retry budget is exhausted. 

The focus of existing evaluation frameworks is mostly accuracy-centric, for example, VQA benchmarks score only answer accuracy, ALCE~\citep{gao2023alce} evaluates citation quality but is text-only, and VLM judges such as LLaVA-Critic~\citep{xiong2025llavacritic} rate responses holistically without attributing failures to specific stages. In contrast, CaVe-VLM-CoT provides fine-grained process supervision by enforcing evidence-grounded reasoning, verifying the faithfulness of intermediate and final outputs, and enabling corrective retrieval when unsupported claims are detected. This design not only improves answer reliability but also increases transparency by exposing where and why failures occur within the retrieval–reasoning–verification pipeline.
   
    
Our contributions are:
(1)\textbf{Step-level citation enforcement with a closed verification loop.} To our knowledge, CaVe-VLM-CoT is the first evaluation framework to jointly enforce inline citation at every reasoning step, validate each citation against its source, and route structured failure signals back to the \textsc{Extractor} for corrective re-querying and re-retrieval.
(2)\textbf{A 23-metric evaluation suite for citation-grounded multimodal reasoning} across four evaluation axes (Extractor, Retriever, Solver, Verifier), anchored by \textbf{CaVeScore}, a composite metric that jointly weights accuracy, citation precision and recall, attribution (AIS), and evidence grounding.
(3)\textbf{Cross-dataset diagnostic analysis.} On ScienceQA ($n{=}1{,}000$), the system achieves $87.1\%$ accuracy and $56.6\%$ CaVeScore; on MMMU ($n{=}500$, 30 subjects), $55.2\%$ and $35.7\%$ respectively. Stage-level analysis identifies retrieval coverage as the primary bottleneck on out-of-domain data.
This paper contributes an architectural framework and evaluation methodology rather than a claim of state-of-the-art accuracy; the empirical results serve as a diagnostic foundation for future work.
\section{Related Work}
\paragraph{Hallucination in VLMs.}
VLMs generate plausible but unfaithful outputs across many tasks~\citep{li2023evaluating, bai2024hallucination}; CHAIR~\citep{rohrbach2018object} and AMBER~\citep{wang2023amber} quantify object and relational hallucinations, and \citet{wu2025generate} report hallucination rates above $40\%$ on knowledge-intensive queries even for instruction-tuned models. The diagnostic literature attributes hallucination to weak visual grounding, incomplete parametric knowledge, or absent external grounding, but none enforce step-wise grounding within a reasoning chain.
\paragraph{CoT for VLMs.}
CoT prompting~\citep{wei2022chain} exposes intermediate steps; LLaVA-CoT~\citep{xu2025llava} adapts this to a four-stage multimodal format, \citet{zhang2025improve} distil GPT-4o traces into smaller models and Critic-V~\citep{zhang2025critic} trains a separate VLM critic. These chains run in a single pass with no per-step citation requirement, and their verification, where it exists, is a passive filter that never routes a failure verdict back into retrieval.
\paragraph{RAG for multimodal reasoning.}
RAG~\citep{lewis2020retrieval} grounds generations in retrieved evidence; ALCE~\citep{gao2023alce} formalises citation-aware generation and \citet{ji2024chain} combine CoT with citation supervision. Multimodal variants RIV-CoT~\citep{corbiere2025retrieval} and MI-RAG~\citep{choi2025multimodal} retrieve evidence for visual CoT but do not verify faithful use of it. Self-RAG~\citep{asai2024selfrag} add self-reflection, but reflection and generation share one forward pass and failure signals never reach the retrieval stage. CaVe-VLM-CoT instead verifies with a dedicated module and propagates failure signals back to the \textsc{Extractor}, surfacing fresh evidence rather than re-reasoning over the same context.
\paragraph{Evaluation of citation quality.}
VQA benchmarks~\citep{antol2015vqa, lu2022scienceqa, marino2019okvqa} report only accuracy; ALCE~\citep{gao2023alce}, FActScore~\citep{min2023factscore}, and AIS-style metrics~\citep{ji2024chain} evaluate citation quality in text only; and multimodal judges LLaVA-Critic~\citep{xiong2025llavacritic}, Prometheus-Vision~\citep{lee2024prometheus}, and VHELM~\citep{lee2024vhelm} score responses holistically without decomposing failures by stage. CaVe-VLM-CoT's 23-metric suite spans evaluation axes (Extractor, Retriever, Solver, Citation Injector, Verifier) and is, to our knowledge, the first to jointly evaluate retrieval quality, step-wise citation faithfulness, and cross-modal grounding within an agentic-RAG pipeline. In sum, prior work neither enforces step-level citation, nor verifies it correctively, nor routes failure signals back to retrieval; CaVe-VLM-CoT closes all three gaps.

\section{Methodology}
\label{sec:methodology}
CaVe-VLM-CoT is built around a single architectural commitment that every factual claim in a reasoning chain must be traceable to a specific piece of retrieved evidence, and any chain that fails this standard must trigger targeted re-retrieval rather than being silently accepted. Figure~\ref{fig:architecture} provides a system overview; full prompt templates and implementation details are in Appendices~\ref{sec:prompts}
and~\ref{sec:implementation}.

\begin{figure*}[h]
    \centering
    \includegraphics[width=\linewidth]{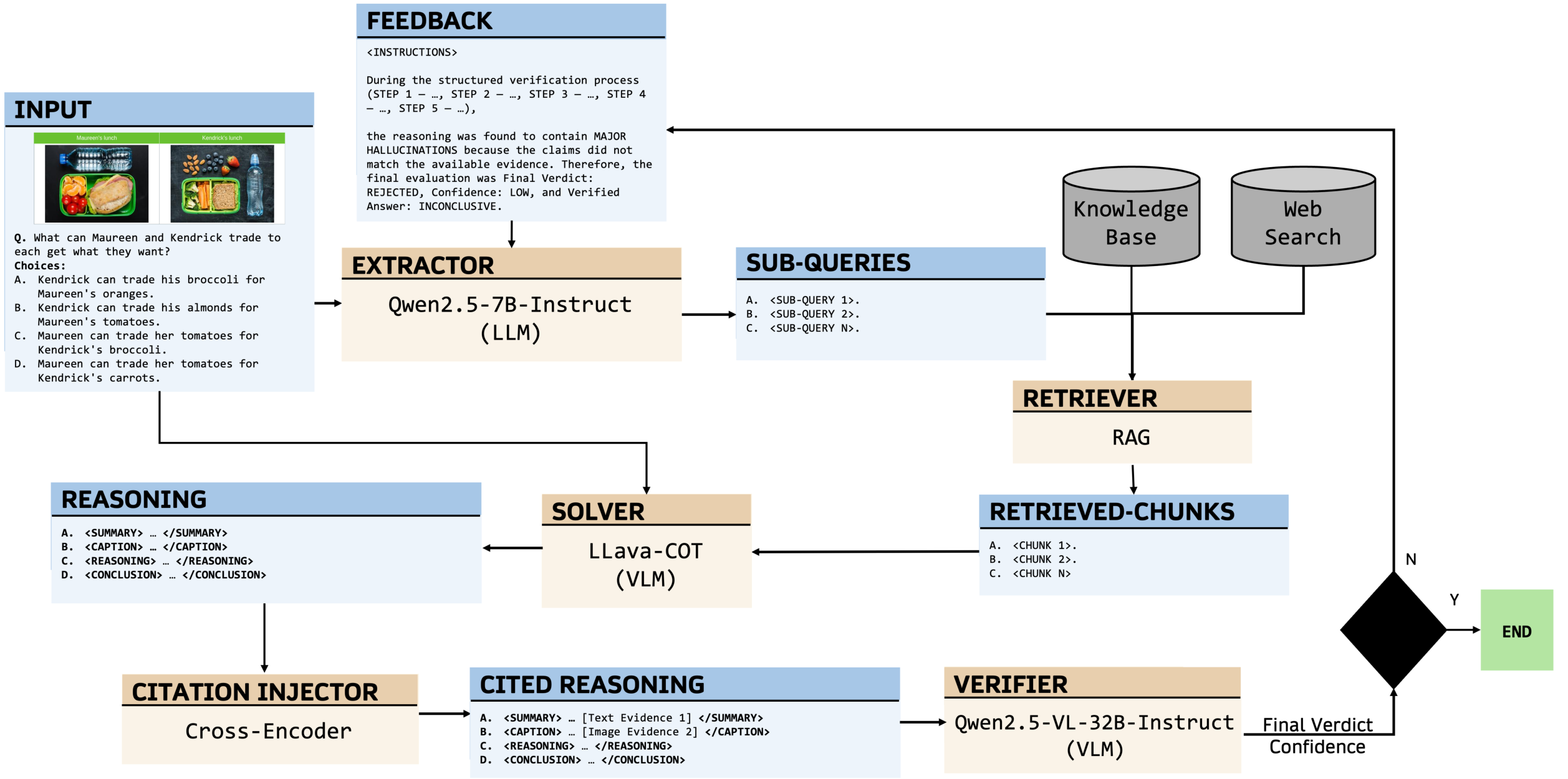}
    \caption{Overview of the CaVe-VLM-CoT pipeline. A multimodal question and its images enter at left. The Extractor (unsloth/Qwen2.5-7B-Instruct-bnb-4bit) decomposes the question into sub-queries routed to a local Knowledge Base and live Web Search; retrieved chunks and input images pass to the Solver (LLaVA-CoT), which produces a structured reasoning trace. The Citation Injector fills uncited claims via cross-encoder matching, and the Verifier (Qwen2.5-VL-32B) audits every citation. On detected hallucinations, structured feedback returns to the Extractor for another retrieval attempt, looping until reasoning is verified or the retry budget is exhausted.}
    \vspace{-0.5em}
    \label{fig:architecture}
\end{figure*}

\subsection{Pipeline Architecture}
\textsc{CaVe-VLM-CoT} is a reflection-based agentic RAG pipeline~\citep{shinn2023reflexion, asai2024selfrag} comprising five sequentially connected modules, \textsc{Extractor}, \textsc{Retriever}, \textsc{Solver}, \textsc{Citation Injector}, and \textsc{Verifier}, with a rule-based feedback loop from the \textsc{Verifier} to the \textsc{Extractor}, executed for up to three attempts ($M{=}3$).

\paragraph{Extractor}
The Extractor decomposes each question into targeted sub-queries that should collectively surface the evidence needed for a correct answer. Hint text, lecture content, and subject labels are withheld, forcing sub-queries to be grounded in the question itself. A few-shot prompt with four in-context examples elicits a structured query list; imperfect outputs are handled by a three-tier fallback parser (Appendix~\ref{subsec:extractor_prompt}). After parsing, a deterministic augmentation step appends one choice-discriminating query per answer option (e.g., \texttt{"sturgeon bottom feeding mouth adaptation"}), guaranteeing retrieval coverage for every
candidate.

On retry attempts, the Extractor additionally receives a structured feedback record from the Verifier specifying the failing claim, the failure type (\texttt{fake citation}, \texttt{misrepresented evidence}, or \texttt{fabricated fact}), what the cited evidence actually says, and a suggested retrieval direction. Previously used queries are listed explicitly to discourage repetition.
Figure~\ref{fig:architecture} illustrates this feedback loop with a concrete example. Full prompt design is described in Appendix~\ref{subsec:extractor_prompt}.

\paragraph{Retriever}
The local knowledge base indexes the ScienceQA lecture corpus using dual FAISS~\citep{douze2024faiss} dense and
BM25~\citep{robertson1994some} sparse indices, fused via Reciprocal Rank Fusion~\citep{cormack2009reciprocal}. For each sub-query, the Retriever executes both local retrieval and web search, including choice-augmented query variants for comparative questions. A cross-encoder reranker scores all pooled candidates, with top local chunks reserved unconditionally to guarantee the knowledge base
coverage. All evidence is keyed to its originating sub-query for
traceability. Corpus preprocessing and indexing details are in
Section~\ref{sec:implementation}.

\paragraph{Solver}
The Solver produces the reasoning trace and final answer in a
structured three- or four-block format (\textsc{Observations}, \textsc{Summary},
\textsc{Reasoning}, \textsc{Conclusion}), making each stage separable for downstream citation injection and verification. Question images are passed directly to the Solver rather than routed through retrieval,
as both sliding-window patch and Grounding-DINO-based~\citep{liu2024grounding}
region proposals degraded performance relative to text-only retrieval in preliminary experiments.

Citation rules are specified negatively: zero citations is preferable to a fabricated one, and domain knowledge claims must be explicitly flagged. A cross-encoder consistency check corrects the stated answer letter when the reasoning text scores more strongly for a different
choice. Full prompt templates are in
Appendix~\ref{subsec:solver_prompt}.

\paragraph{Citation Injector}
The Citation Injector operates as a post-processing step between the Solver and the Verifier, filling citation gaps in the reasoning trace by matching uncited claims against
the retrieved evidence via cross-encoder scoring. Claims within the visual observations block are excluded from text-evidence matching and handled separately via a pattern-based
image citation strategy. A citation is injected when the cross-encoder score exceeds a tuned threshold $\tau=0.4$. Critically, the injector replicates the exact evidence
numbering used in the Solver's prompt, guaranteeing that any injected label refers to the same chunk the Solver saw, preventing the citation misalignment that would otherwise cause the Verifier to flag valid claims as fabricated.

\paragraph{Verifier}
The Verifier receives the complete reasoning trace, all retrieved evidence, and the original images, and performs claim-level fact-checking. For each cited claim, it checks whether the referenced evidence supports the stated fact, whether any citation index is out of range, and whether any image description contradicts the visual
content. It produces a structured verdict with hallucination severity (none, minor, major) and confidence (high, medium, low). Each detected hallucination generates a feedback record injected into the Extractor's prompt on the next attempt
(Appendix~\ref{subsec:verifier_prompt}).

\subsection{Feedback Loop and Retry Logic}
Major hallucinations always trigger a retry; minor hallucinations trigger a retry only when verifier confidence is medium or low. A verified verdict or any verdict after $M{=}3$ attempts terminates the pipeline. On retry, the Extractor receives the feedback note,
previously used queries, and full attempt history, ensuring the
evidence base changes materially on each cycle.

\section{Experimental Setup}
\label{sec:experiments}
\paragraph{Implementation.}
All configurations share identical hardware (3$\times$NVIDIA A40 48\,GB), tokenizers, and generation hyperparameters (Solver $T{=}0.3$, Verifier $T{=}0.0$, max 1024/2048 new tokens). Retrieval uses BM25 + FAISS dense retrieval with \texttt{all-MiniLM-L6-v2}, cross-encoder re-ranking, $k{=}5$ chunks per sub-query, and up to 8 sub-queries per question, held fixed across configurations.

\paragraph{Datasets.}
We evaluate on a stratified subsample of ScienceQA~\citep{lu2022scienceqa} ($n{=}1{,}000$; $52.3\%$ with images) spanning natural, social, and language science, and on MMMU~\citep{yue2024mmmu} ($n{=}500$, 30 subjects across \texttt{dev}/\texttt{validation}) for generalization. The ScienceQA budget keeps the 95\% Wilson CI for accuracy within $\pm 3.1$\,pp and gives $>80\%$ power to detect a 6\,pp gap between configurations at $\alpha{=}0.05$, matching our ablation resolution. MMMU preserves $\geq 16$ items per subject. Details in Appendix~\ref{sec:implementation}.
\paragraph{Metrics.}
\label{subsec:metrics_main}
No single existing metric jointly rewards correct, well-cited, attributable, and evidence-grounded answers. CaVeScore combines five complementary signals:
\begin{multline}
  \text{CaVeScore} = w_1 \!\cdot\! \text{Acc} + w_2 \!\cdot\! \text{CitePrec} + w_3 \!\cdot\! \text{CiteRec} \\
  + w_4 \!\cdot\! \text{AIS} + w_5 \!\cdot\! \text{Grounding}
  \label{eq:cavescore_main}
\end{multline}
where \textbf{Acc} is exact-match correctness; \textbf{CitePrec} is mean NLI entailment between each cited chunk and its claim; \textbf{CiteRec} is the fraction of factual sentences carrying $\geq 1$ citation; \textbf{AIS} is the fraction of reasoning steps entailed by retrieved evidence; and \textbf{Grounding} is the maximum entailment between cited evidence and the final answer. We set $(w_1,\ldots,w_5){=}(0.4,0.2,0.2,0.1,0.1)$, prioritising correctness while requiring claims be both supported and cited. Weight sensitivity is in \S\ref{app:sensitivity}; formal definitions of all 23 metrics are in Appendix~\ref{sec:metrics}.

\paragraph{Models.}
The pipeline uses five models. The \textbf{Extractor} (Qwen2.5-7B-Instruct~\citep{qwen25}, 4-bit NF4 quantised) generates retrieval sub-queries. The \textbf{Retriever} combines dense embeddings from \texttt{all-MiniLM-L6-v2}~\citep{reimers2019sentence} with cross-encoder re-ranking via \texttt{ms-marco-MiniLM-L6-v2}. The \textbf{Solver}, the VLM that produces the chain-of-thought answer over images and evidence, is LLaVA-CoT~\citep{xu2025llava}. The \textbf{Citation Injector} reuses the \texttt{ms-marco} reranker to attach citations to uncited claims. The \textbf{Verifier}, the largest model in the stack, is Qwen2.5-VL-32B-Instruct~\citep{bai2025qwen25vl}; its scale is deliberate, since \S\ref{subsec:discussion} shows the feedback loop's value is bottlenecked by verifier capacity. At evaluation time only, \texttt{cross-encoder/nli-deberta-v3-base}~\citep{he2021debertav3} scores entailment for CitePrec, AIS, and Grounding. Full hyperparameters in Appendix~\ref{sec:implementation}.

\paragraph{Ablation Configurations.}
\label{subsec:configs}
We test whether each pipeline stage contributes measurably to CaVeScore with three configurations that progressively add components (Table ~\ref{tab:configs}).

\begin{table}[h]
\centering
\resizebox{\columnwidth}{!}{%
\begin{tabular}{lccccc}
\toprule
\textbf{Config.} & \textbf{Ext.} & \textbf{Ret.} & \textbf{Sol.} & \textbf{Cit.I.} & \textbf{Ver.} \\
\midrule
\sys{Solver-Only}   & $\varnothing$ & $\varnothing$ & \checkmark & $\varnothing$ & $\varnothing$ \\
\sys{Ret.-Solver}   & \checkmark & \checkmark & \checkmark & $\varnothing$ & $\varnothing$ \\
\rowcolor{bestrow}
\sys{CaVe-VLM-CoT}  & \checkmark & \checkmark & \checkmark & \checkmark & \checkmark \\
\bottomrule
\end{tabular}
}
\caption{Ablation configurations. \checkmark\,=\,active; $\varnothing$\,=\,removed.}
\label{tab:configs}
\end{table}

\begin{table*}[t]
\centering
\renewcommand{\arraystretch}{1.5}
\setlength{\tabcolsep}{4pt}
\resizebox{\textwidth}{!}{%
\begin{tabular}{@{}cl lll @{\hspace{1.2em}} cl lll@{}}
\toprule
& \textbf{Metric} & \textbf{ScienceQA} & \textbf{MMMU} & $\boldsymbol{\Delta}$
& & \textbf{Metric} & \textbf{ScienceQA} & \textbf{MMMU} & $\boldsymbol{\Delta}$ \\
\midrule
\multirow{4}{*}{\rotatebox[origin=c]{90}{\textit{\color{steelblue}Extractor}}}
  & Coverage       & $0.682 \pm 0.201$ & $0.501 \pm 0.203$ & $-0.181$
& \multirow{11}{*}{\rotatebox[origin=c]{90}{\textit{\color{steelblue}Solver}}}
  & Accuracy          & $0.871 \pm 0.335$ & $0.552 \pm 0.498$ & $-0.319$ \\
  & Specificity    & $0.999 \pm 0.006$ & $0.987 \pm 0.111$ & $-0.012$
& & Text Cite Prec.   & $0.322 \pm 0.375$ & $0.120 \pm 0.241$ & $-0.202$ \\
  & Hit Rate       & $0.304 \pm 0.460$ & $0.082 \pm 0.256$ & $-0.222$
& & QI Cite Prec.     & $0.238 \pm 0.252$ & $0.280 \pm 0.371$ & $+0.042$ \\
  & Subquery Count & $7.491 \pm 0.715$ & $7.632 \pm 1.169$ & $+0.141$
& & QI Cite Coverage  & $0.713 \pm 0.499$ & $0.282 \pm 0.372$ & $-0.431$ \\
\cmidrule(r){1-5}
\multirow{4}{*}{\rotatebox[origin=c]{90}{\textit{\color{steelblue}Retriever}}}
  & Recall@2       & $0.518 \pm 0.277$ & $0.085 \pm 0.176$ & $-0.433$
& & Citation Precision & $0.543 \pm 0.399$ & $0.325 \pm 0.358$ & $-0.218$ \\
  & Precision@2    & $0.150 \pm 0.273$ & $0.050 \pm 0.159$ & $-0.100$
& & Citation Recall   & $0.241 \pm 0.235$ & $0.197 \pm 0.320$ & $-0.044$ \\
  & MRR            & $0.126 \pm 0.230$ & $0.027 \pm 0.103$ & $-0.099$
& & AIS               & $0.408 \pm 0.304$ & $0.153 \pm 0.201$ & $-0.255$ \\
  & NDCG@2         & $0.123 \pm 0.228$ & $0.027 \pm 0.103$ & $-0.096$
& & Hallucination Rate & $0.592 \pm 0.304$ & $0.835 \pm 0.224$ & $+0.243$ \\
\cmidrule(r){1-5}
\multirow{4}{*}{\rotatebox[origin=c]{90}{\textit{\color{steelblue}Verifier}}}
  & Decision Correct       & $0.525 \pm 0.500$ & $0.670 \pm 0.471$ & $+0.145$
& & Grounding Score   & $0.197 \pm 0.343$ & $0.163 \pm 0.296$ & $-0.034$ \\
  & Halluc.\ Det.\ Correct & $0.309 \pm 0.462$ & $0.234 \pm 0.424$ & $-0.075$
& & Is Grounded       & $0.220 \pm 0.413$ & $0.180 \pm 0.381$ & $-0.040$ \\
  & Confidence Approp.     & $0.494 \pm 0.500$ & $0.460 \pm 0.499$ & $-0.034$
& & \cellcolor{bestrow}\textbf{CaVeScore} & \cellcolor{bestrow}$\mathbf{0.566 \pm 0.221}$ & \cellcolor{bestrow}$\mathbf{0.357 \pm 0.256}$ & \cellcolor{bestrow}$\mathbf{-0.209}$ \\
  & Feedback Quality       & $0.274 \pm 0.446$ & $0.329 \pm 0.470$ & $+0.055$
& & & & & \\
\bottomrule
\end{tabular}
    }
\caption{Full metric summary on ScienceQA ($n{=}1{,}000$) and MMMU ($n{=}500$, 30 subjects), reported as mean $\pm$ std.
For binary indicators (e.g.\ Accuracy, Hit Rate, Is Grounded), std is the Bernoulli standard deviation $\sqrt{p(1{-}p)}$ across the sample; continuous metrics (CitePrec, AIS, CaVeScore) report the empirical standard deviation. $\Delta$ = MMMU $-$ ScienceQA.}
\label{tab:full_results}
\end{table*}

\begin{table*}[h]
\centering
\setlength{\tabcolsep}{4pt}
\renewcommand{\arraystretch}{1.15}
\begin{tabular}{lccccc}
\toprule
\textbf{Configuration} & \textbf{Acc.} & \textbf{AIS} & \textbf{CitePrec} & \textbf{CaVe} & $\boldsymbol{\Delta}$\textbf{CaVe} \\
\midrule
\sys{Solver-Only}        & 0.730 & 0.125 & 0.000 & 0.314 & --- \\
\sys{Retriever-Solver}   & 0.704 & 0.318 & 0.249 & 0.404 & \up{0.090} \\
\rowcolor{bestrow}
\sys{CaVe-VLM-CoT}       & \tblbest{0.871} & \tblbest{0.408} & \tblbest{0.543} & \tblbest{0.566} & \tblbest{\up{0.252}} \\
\bottomrule
\end{tabular}
\caption{Ablation results on ScienceQA ($n{=}1{,}000$). \textbf{Bold} = best per column. CaVeScore is the primary ranking criterion. $\Delta$CaVe is the gain over \sys{Solver-Only}.}
\label{tab:ablation}
\end{table*}

\sys{Solver-Only} receives the question, choices, and images but no retrieved evidence, generating CoT from parametric knowledge alone. \sys{Retriever-Solver} adds hybrid retrieval and web search but omits the Citation Injector, Verifier, and feedback loop, isolating the contribution of the verification machinery. \sys{CaVe-VLM-CoT} is the full pipeline.

\section{Results}
\label{subsec:results}
The results are discussed below:
\subsection{Overall Results}
Table~\ref{tab:full_results} reports all metrics on both datasets. The central diagnostic is the gap between accuracy and grounding: on ScienceQA, accuracy ($87.1\%$) far exceeds AIS ($40.8\%$), meaning a substantial share of correct answers rely on parametric knowledge rather than retrieved evidence. On MMMU, this decoupling intensifies; retrieval and grounding metrics (Recall@2, AIS, CitePrec) collapse far more steeply than accuracy ($\Delta$ column), the signature of a corpus-coverage failure rather than a reasoning failure. The VLM generalises across domains; the retrieval corpus does not.

\subsection{Cross-Dataset Generalisation}
\label{subsec:generalization}

The pipeline was applied to MMMU without retraining, prompt changes, or architectural modifications. Table~\ref{tab:full_results} decomposes the cross-dataset gap by stage. Retrieval collapses (Hit Rate $-22.2$\,pp, Recall@2 $-43.3$\,pp), reflecting that the ScienceQA lecture corpus contains little content relevant to MMMU's 30 university-level disciplines (e.g., clinical medicine, mechanical engineering, art history). With retrieval effectively absent, the Solver falls back on parametric knowledge, consistent with the elevated hallucination rate ($+24.3$\,pp). Yet accuracy degrades only $-31.9$\,pp and stays well above the random baseline,
 showing the underlying VLM still contributes useful parametric knowledge when retrieval fails.

The pattern across metrics is itself diagnostic. The retrieval and grounding metrics (Recall@2, AIS, CitePrec) collapse much further than accuracy, the signature of a corpus-coverage failure rather than a reasoning failure. A response-level metric would surface a single ``MMMU accuracy drop'' number; the stage-level decomposition identifies knowledge-base coverage as the bottleneck, not query formulation or reasoning architecture, and points directly to corpus expansion as the actionable fix.

\subsection{Ablations}

Table~\ref{tab:ablation} reports the four headline metrics across the three configurations on the ScienceQA sample.

\noindent Moving from \sys{Solver-Only} to \sys{Retriever-Solver} reduces accuracy by $2.6$ points (the retriever occasionally surfaces distracting evidence) but adds $+19.3$ points of AIS and $+24.9$ of citation precision: retrieval converts parametric claims into attributable, evidence-backed ones, driving CaVeScore up by $+9.0$ points. Moving from \sys{Retriever-Solver} to the full \sys{CaVe-VLM-CoT} adds another $+16.7$ points of accuracy, $+9.0$ of AIS, and $+29.4$ of citation precision, for a further $+16.2$ CaVeScore gain. The verification loop not only improves grounding (as expected) but also substantially lifts accuracy, because the calibrated 32B Verifier issues more reliable rejection signals that drive targeted re-retrieval and surface evidence missed on the first pass.

\subsection{Discussion}
\label{subsec:discussion}

The single most consequential design choice is the scale of the Verifier (Table~\ref{tab:verifier_scale}). Under the 8B model (Qwen2.5-VL-8B), Decision Correctness was $9.9\%$, well below the $50\%$ chance rate for a binary verdict, and Feedback Quality was $0\%$ on every evaluated sample: the feedback loop was architecturally sound but empirically inert. At 32B, Decision Correctness rises to $52.5\%$ and Feedback Quality to $27.4\%$, and the accuracy gain from adding the Verifier over \sys{Retriever-Solver} jumps from $+4.3$ to $+16.7$ points.
This confirms that the loop's value was previously bottlenecked by the quality-control capacity of the smaller model, not by the loop's design: a verifier that cannot reliably tell grounded from ungrounded reasoning produces feedback that cannot steer re-retrieval.

\begin{table}[h]
\centering
\setlength{\tabcolsep}{5pt}
\renewcommand{\arraystretch}{1.15}
\begin{tabular}{lcc}
\toprule
\textbf{Metric} & \textbf{8B} & \textbf{32B} \\
\midrule
Decision Correctness        & $9.9\%$   & $52.5\%$ \\
Feedback Quality            & $0.0\%$   & $27.4\%$ \\
\midrule
$\Delta$Acc vs.\ Ret.-Solver & $+4.3$\,pp & $+16.7$\,pp \\
\bottomrule
\end{tabular}
\caption{Impact of Verifier scale on verification quality and upstream accuracy. $\Delta$Acc is the accuracy gain over \sys{Retriever-Solver} ($70.4\%$).}
\label{tab:verifier_scale}
\end{table}

\paragraph{Residual failure modes.}
Two bottlenecks remain well characterised. First, verifier calibration is improved but not complete: Decision Correctness is $52.5\%$ (just above chance). The dominant residual failure mode is INCONCLUSIVE rejection handling, where the Verifier flags a problem but cannot name the correct answer, overriding a Solver that was in fact right. Second, retrieval coverage on out-of-domain data is thin: MMMU's $8.2\%$ Hit Rate is the mechanism through which the cross-dataset accuracy gap appears. Both are concrete, stage-identified failure modes that response-level accuracy alone cannot expose, and both point to actionable fixes (verifier scale or fine-tuning, and corpus expansion, respectively).

Ablation ordering is preserved under all six alternative CaVeScore weightings ($\Delta$ range $+0.20$ to $+0.30$; full table in Appendix~\ref{app:sensitivity}).
\section{Conclusion}
\label{sec:conclusion}
We introduced CaVe-VLM-CoT, a five-stage reflection-based agentic-RAG framework built on the premise that hallucination is a grounding failure, not a generation defect, and must be caught structurally rather than mitigated through prompting alone. By tying every reasoning step to a verifiable citation and routing structured failure signals back to the \textsc{Extractor} when that link breaks, it converts verification from a passive post-hoc filter into an active corrective mechanism.

The results both validate this commitment and sharpen its limits. The pipeline transfers to MMMU's college-level reasoning across 30 disciplines with no architectural or prompt changes, yet stage-level metrics expose a sharp retrieval-coverage drop (Hit Rate $30.4\%\rightarrow 8.2\%$) that response-level accuracy would conceal, and ablations confirm each stage contributes measurably to CaVeScore under all six alternative weightings. The hallucination rate ($59.2\%$ on ScienceQA, $83.5\%$ on MMMU) and residual verifier miscalibration reveal that correctness and grounding remain partially decoupled: the system reaches right answers for partially wrong reasons, a failure mode aggregate accuracy cannot expose. To our knowledge, CaVe-VLM-CoT is the first framework to make verifier calibration measurable at this granularity and expose it as a primary bottleneck in citation-grounded multimodal reasoning. 

\section{Future Work}
\label{sec:future_work}
The evaluation points to three directions. First, \textbf{Verifier rejection precision}: at 32B, Decision Correctness reaches $52.5\%$ (from $9.9\%$), but a REJECTED verdict with \textsc{Verified Answer: INCONCLUSIVE} still cannot reliably override a correct Solver; recovering the Solver's answer in these cases, rather than defaulting to rejection, would remove most residual false rejections without retraining. Whether the remaining gap is a matter of scale or of how verification is supervised is open. Second, \textbf{the visual retrieval gap}: image evidence does not yet carry the citation-grounding guarantees of retrieved text. Region-proposal models for citation-relevant crops and cross-modal rerankers that score image-claim pairs as reliably as text could close it. Third, \textbf{domain coverage}, the dominant bottleneck for cross-domain transfer: extending beyond multiple-choice QA to open-ended generation, longer chains, and domain-specific corpora (e.g., medical imaging, legal analysis) would test whether the architectural commitments hold where failure modes and evidence types differ.

\clearpage
\section{Limitations}
\label{sec:limitations}

Several limitations should be acknowledged. First, the Verifier remains the binding constraint on the feedback loop. Upgrading to Qwen2.5-VL-32B substantially improves calibration improving Decision Correctness from 9.9\%
to 52.5\% and Feedback Quality from 0\% to 27.4\% but both remain below ideal, and Hallucination Detection Correctness is only 30.9\%. The dominant failure mode persists: the Verifier flags a problem but cannot name the correct answer, occasionally overriding a Solver that was in fact correct. Further scale or task-specific fine-tuning would likely help.

Second, image-based retrieval remains unsolved. Both sliding-window patch retrieval and grounding-DINO region proposals degraded performance relative to text-only retrieval, forcing us to pass question images directly to the Solver. Visual evidence therefore cannot carry the same citation-grounding guarantees as retrieved text, leaving a modality gap in the attribution framework.

Third, both evaluation datasets used in our experiments (ScienceQA and MMMU) are multiple-choice. Whether these results generalise to open-ended generation, domain-specific corpora with proprietary knowledge bases (e.g., clinical or legal reasoning), or substantially longer reasoning chains remains open. Relatedly, the CaVeScore weights encode a specific design priority; our sensitivity analysis (Section~\ref{app:sensitivity}) confirms ablation rankings are stable across seven configurations, though optimal weighting may vary by domain.

\section{Broader Impact Statement}
\label{sec:impact}

CaVe-VLM-CoT aims to make vision-language reasoning more transparent and verifiable by enforcing citation grounding at every reasoning step. In knowledge-intensive domains such as scientific education, medical imaging, and legal analysis, tracing each claim to a specific piece of evidence is a prerequisite for responsible deployment. By surfacing hallucination at the stage level rather than hiding it behind aggregate accuracy, the framework gives practitioners a way to locate and address failure modes before they reach end users.
Some risks remain. A well-cited answer is not necessarily a correct one, and citation markers may lead users to over-trust outputs without checking the cited sources themselves. The evaluation suite, though more informative than accuracy alone, inherits the limitations of the component models it relies on, so its scores should be read as diagnostic signals rather than guarantees. The system also depends on web search for evidence, which exposes it to misinformation or bias in external sources.
We release the evaluation framework and metric definitions so the community can apply stage-level diagnostic evaluation to other multimodal reasoning systems, and we encourage future work to stress-test CaVeScore across diverse domains and failure modes.

\section*{Use of Large Language Models}
\label{sec:llm_disclosure}

In accordance with transparency requirements, we disclose the 
use of large language models during the preparation of this 
manuscript. Claude (Anthropic) was used as a writing aid for 
drafting, paraphrasing, and editing prose in the manuscript 
text. All technical content, experimental design, system 
architecture, evaluation framework, metric definitions, and 
reported results are the original intellectual contribution 
of the authors. The authors reviewed and edited all 
LLM-assisted text and take full responsibility for the 
content of this publication.

\bibliography{references}

\newpage
\appendix
\section*{Appendix}
\setcounter{section}{0}
\setcounter{subsection}{0}
\setcounter{subsubsection}{0}

\renewcommand{\thesection}{\Alph{section}}
\renewcommand{\thesubsection}{\thesection.\arabic{subsection}}
\renewcommand{\thesubsubsection}{\thesubsection.\arabic{subsubsection}}

\setcounter{figure}{0}
\setcounter{table}{0}

\renewcommand{\thefigure}{\thesection\arabic{figure}}
\renewcommand{\thetable}{\thesection\arabic{table}}

\section{Methodology Details}
\subsection{Corpus Construction}

Our primary evaluation uses ScienceQA~\citep{lu2022scienceqa}, a large-scale multimodal benchmark spanning natural science, social science, and language science questions, each paired with multiple-choice answer options, optional context images, lecture text, and annotated solutions. As part of preprocessing, images are normalised to a uniform spatial representation of 224x224; a pre-trained captioning model~\citep{li2022blip} and optical character recognition (OCR) are applied to produce textualised descriptions of each image's content; and subject, topic, and skill metadata are stored alongside each example. A critical design choice is that lecture text, hints, and subject metadata are \textit{not} injected into the Extractor prompt; they reside in the knowledge base and must be surfaced by well-formed retrieval queries. If this metadata were provided directly to the Extractor, the system could trivially match against it without demonstrating genuine retrieval capability, artificially inflating Hit Rate and Recall metrics. By withholding it, we ensure the pipeline is evaluated under realistic inference-time conditions where such annotations are unavailable. Full preprocessing details are in Appendix~\ref{sec:implementation}.

\subsection{Knowledge Base Construction}
The local knowledge base is built from the ScienceQA lecture corpus and indexed using two complementary indices maintained in parallel:

\begin{itemize}[noitemsep, topsep=4pt]
  \item A \textbf{dense vector index} (FAISS \citep{douze2024faiss}) that encodes text chunks into fixed-length embeddings for semantic similarity search, capturing 
  meaning-level matches even when surface words differ.
  \item A \textbf{sparse term-frequency index} (BM25 \citep{robertson1994some}) for keyword 
  matching, retrieving chunks that share exact terms with the query 
  but may lie far apart in embedding space.
\end{itemize}
\noindent This dual-index design captures evidence that either 
approach would miss in isolation. The two indices are fused at 
retrieval time via the Reciprocal Rank 
Fusion~\citep{cormack2009reciprocal}, so the fusion strategy can 
be tuned independently of index construction.

\section{Prompt Engineering}
\label{sec:prompts}

\subsection{Extractor Prompt}
\label{subsec:extractor_prompt}

The Extractor receives only the question, answer choices, and image-derived context (BLIP
captions and OCR text). No lecture text, hints, or subject labels are provided. Its task is
to output a structured list of 3-12 word retrieval queries covering three complementary
families, \textit{(i)} definitional and conceptual queries about key terms in the question;
\textit{(ii)} choice-discriminating queries that help surface evidence distinguishing among
the answer options; and \textit{(iii)} comparative queries that relate multiple choices
against each other.

The prompt terminates with an open bracket to prime structured list generation. Four
in-context examples cover distinct reasoning types: geographic comparison, grammar
classification, physical science, and title capitalisation, each demonstrating a query set
that spans at least one definitional query and one query per answer choice. Output parsing
operates in three fallback tiers, full structured list parsing (primary path), partial
recovery from truncated output (secondary), and line-by-line extraction from plain
numbered-list output (last resort). Parsed queries are validated for length (2-12 words),
content-bearing tokens, and non-repetition of the question stem. A deterministic
choice-discriminating augmentation step then appends one query per answer choice,
guaranteeing retrieval coverage for every candidate regardless of model output. On retry
iterations, structured Verifier feedback is spliced between the question block and the
opening bracket, with previously used queries listed explicitly to discourage repetition.

\begin{lstlisting}[caption={Extractor prompt skeleton (abbreviated).}]
Generate search queries to retrieve the evidence needed to answer a
multiple-choice question. Your queries will be run against a knowledge
base AND a web search engine.

OUTPUT FORMAT: a structured list of query strings, nothing else.
Each query: 3-12 words, plain keywords or short phrases, no question marks.

[Four few-shot examples omitted for brevity]

Question: {question}
Choices:  {choices}
{image_context}[
\end{lstlisting}

\subsection{Solver Prompts}
\label{subsec:solver_prompt}

The Solver maintains two prompt templates selected at runtime based on whether the question
has associated images. Both enforce a rigid four-block output format.

\paragraph{Text-only questions.} The model structures its response as:
\begin{enumerate}[noitemsep, topsep=2pt]
  \item \texttt{<SUMMARY>} states the core problem without factual claims;
  \item \texttt{<REASONING>} each step must follow \textit{``According to
        [Text Evidence N], \ldots''} or \textit{``From domain knowledge, \ldots''};
  \item \texttt{<CONCLUSION>} restates the answer without introducing new facts.
\end{enumerate}

\paragraph{Vision questions.} An \texttt{<OBSERVATIONS>} section is prepended, requiring
the model to describe each image using mandatory \texttt{[Question Image N]} citations
before reasoning begins. A cross-encoder consistency check runs post-parse: for each answer
choice the cross-encoder scores the pair (reasoning text, candidate answer), and if the
highest-scoring choice differs from the stated answer by more than 0.5 score units, the
answer is silently corrected. This guards against a failure mode common in smaller VLMs
where the model reasons to the correct conclusion but writes the wrong letter.

\paragraph{Citation rules.} The model is instructed explicitly and negatively: it is
strictly better to have zero citations than to cite evidence that does not support a claim.
Domain knowledge claims must be prefixed with \textit{``From domain knowledge, \ldots''}
only after consulting all retrieved evidence. A low-temperature retry ($T=0.1$) fires if
the initial generation lacks text citations despite substantive retrieved evidence, or lacks
visual observations despite images being present.

\subsection{Verifier Prompt}
\label{subsec:verifier_prompt}

The Verifier is designed as a strict fact-checker operating over the complete reasoning
trace, the full retrieved evidence set (formatted with the same citation labels as the
Solver's prompt), and the original question images. Its task is claim-level, not
response-level. For each cited claim it checks whether the referenced evidence chunk
actually supports the stated fact, whether any citation index is out of range, and whether
any image description contradicts what is visually present. Common knowledge inferences,
paraphrases of the question, and partial evidence support are explicitly listed as
non-hallucinations to prevent over-flagging of legitimate reasoning.

The output format is strictly structured and parsed via regex:

\begin{lstlisting}[caption={Verifier structured output format.}]
Hallucination Check: [NONE DETECTED | MINOR HALLUCINATIONS | MAJOR HALLUCINATIONS]

[If hallucinations detected:]
Claim:    "<exact quote from solver>"
Issue:    fake citation | not in evidence | misrepresented | fabricated
Evidence: <what the evidence actually says, or "doesn't exist">

Final Verdict:   [VERIFIED | REJECTED]
Confidence:      [HIGH | MEDIUM | LOW]
Verified Answer: [A/B/C/D/E | INCONCLUSIVE]
\end{lstlisting}

\noindent A critical alignment requirement: the Verifier reconstructs the evidence index
using identical filtering and capping logic to the Solver, capping at five chunks per
sub-query and ten total. If these rules differ by even one entry, a citation valid in the
Solver's prompt resolves to a different chunk in the Verifier's context, causing correct
citations to be flagged as fabricated and triggering spurious retry cycles.

\begin{table*}[h]
\centering
\small
\setlength{\tabcolsep}{4pt}
\renewcommand{\arraystretch}{1.15}
\begin{tabular}{llll}
\toprule
\textbf{Stage} & \textbf{Model} & \textbf{Size} & \textbf{Role} \\
\midrule
Extractor          & Qwen2.5-7B-Instruct~\citep{qwen25}            & 7B (4-bit) & Sub-query generation \\
Retriever          & all-MiniLM-L6-v2~\citep{reimers2019sentence}  & 22M        & Dense embedding \\
Re-ranker          & ms-marco-MiniLM-L-6-v2                        & 22M        & Cross-encoder rerank \\
Solver             & LLaVA-CoT~\citep{xu2025llava}                 & 11B        & VLM reasoning + CoT \\
Citation Injector  & ms-marco-MiniLM-L-6-v2                        & 22M        & Citation attachment \\
Verifier           & Qwen2.5-VL-32B-Instruct~\citep{bai2025qwen25vl} & 32B      & Verdict + feedback \\
\midrule
\multicolumn{4}{l}{\textit{Evaluation only (not in pipeline):}} \\
NLI scorer         & nli-deberta-v3-base~\citep{he2021debertav3}   & 184M       & CitePrec, AIS, Grounding \\
\bottomrule
\end{tabular}
\caption{Models used in each pipeline stage. Sizes approximate; all hyperparameters in Appendix~\ref{sec:implementation}.}
\label{tab:models}
\end{table*}

\section{Evaluation Framework}
\label{sec:metrics}

\paragraph{Metrics.}
For the primary results we foreground four headline metrics:
\begin{itemize}[noitemsep, topsep=2pt]
  \item \textbf{Accuracy}: exact match of the predicted answer letter $\hat{y}$ to the gold index $y^*$, averaged over $N$ examples:
  $\text{Accuracy} = \tfrac{1}{N}\sum_{i=1}^{N} \mathbb{1}\{\hat{y}_i = y^*_i\}$.
  \item \textbf{AIS}: fraction of factual reasoning sentences $s \in S$ for which at least one retrieved evidence chunk $e \in \mathcal{E}$ entails $s$ under an NLI model:
  $\text{AIS} = \tfrac{1}{|S|} \sum_{s \in S} \mathbb{1}\{\max_{e \in \mathcal{E}} P_{\text{NLI}}(\text{entail} \mid e, s) > \tau_{\text{AIS}}\}$, with $\tau_{\text{AIS}}=0.35$.
  \item \textbf{Citation Precision}: mean NLI entailment score of each (evidence, claim) citation pair in the Solver's response:
  $\text{CitePrec} = \tfrac{1}{|\mathcal{C}|}\sum_{(e,c)\in \mathcal{C}} \text{score}_{\text{NLI}}(e, c)$, where $\text{score}_{\text{NLI}} \in \{0, 0.5, 1\}$ for contradiction/neutral/entailment.
  \item \textbf{CaVeScore}: a weighted composite of Accuracy, Citation Precision, Citation Recall, AIS, and Grounding.
\end{itemize}
Accuracy alone combines correct answers achieved through hallucinated reasoning with those that are genuinely grounded. CaVeScore is therefore our primary ranking criterion, as it jointly rewards correctness \emph{and} transparent attribution.

CaVeScore differs from prior evaluation approaches in that it jointly 
measures answer correctness, citation precision and recall, step-level 
attribution (AIS), and evidence grounding within a single composite. 
Accuracy and BLEU/ROUGE capture only the first dimension; ALCE and AIS 
target citation and attribution respectively but do not span both; and 
holistic VLM judges such as LLaVA-Critic~\citep{xiong2025llavacritic} 
and VHELM~\citep{lee2024vhelm} score responses as a whole rather than 
decomposing quality across these axes. A full dimension-by-dimension 
comparison is provided in Appendix~\ref{app:metric_comparison}.
We evaluate \textsc{CaVe-VLM-CoT} along four axes, Extractor quality, Retriever quality, Solver quality, and Verifier quality. Throughout this section, let $Q$ denote the set of sub-queries generated by the Extractor, $\phi(\cdot)$ an embedding function (\texttt{all-MiniLM-L6-v2}), $\text{top}_K(q)$ the $K$ highest-ranked chunks returned for sub-query $q$, $y^*$ the gold answer index, and $\hat{y}$ the predicted answer letter. $P_{\text{NLI}}(\text{entailment} \mid p, h)$ is the entailment probability from a natural language inference model with premise $p$ and hypothesis $h$. $\mathcal{E}$ denotes the full set of retrieved evidence chunks, and $\mathcal{E}_\text{cited} \subseteq \mathcal{E}$ the subset explicitly cited in the Solver's response. $\mathbb{1}\{\cdot\}$ is the indicator function.

\subsection{Extractor Metrics}

\paragraph{Coverage Score.}
Measures how thoroughly the generated sub-queries cover the key terms in the input. Let $\mathcal{T}$ be the set of non-stopword alphabetic tokens (length $\geq 3$) extracted from the question and answer choices, and $S$ the concatenated text of all sub-queries:
\begin{equation}
  \text{Coverage} = \frac{|\{t \in \mathcal{T} : t \in S\}|}{|\mathcal{T}|}
  \label{eq:coverage}
\end{equation}
A score of 1.0 indicates every key term appears in at least one sub-query.

\paragraph{Specificity Score.}
Quantifies whether individual sub-queries are precise enough to retrieve targeted evidence. For each sub-query $q$ with word count $w$, a length score $\ell(q) \in \{0.3, 0.6, 1.0\}$ is assigned based on thresholds ($\ell = 0.3$ if $w < 3$; $\ell = 1.0$ if $w > 15$; $\ell = 0.6$ otherwise), with a fixed penalty of 0.2 subtracted when $q$ begins with a generic preamble (e.g., ``What is''):

\scriptsize
\begin{equation}
  \text{Specificity} = \frac{1}{|Q|} \sum_{q \in Q} \max\!\bigl(0,\; \ell(q) - \text{penalty}(q)\bigr)
  \label{eq:specificity}
\end{equation}

\normalsize
\paragraph{Hit Rate.}
A binary indicator that fires when any retrieved chunk, across all sub-queries, is semantically close to the gold answer. Specifically, it requires at least one top-$K$ chunk $c$ for some sub-query $q$ to exceed a cosine similarity threshold $\theta = 0.5$ with the embedding of the concatenated question and gold answer:

\scriptsize
\begin{equation}
  \text{Hit} = \bigvee_{q \in Q}\; \bigvee_{c \in \text{top}_K(q)} \mathbb{1}\!\left\{\cos\!\bigl(\phi(q + \text{gold}),\, \phi(c)\bigr) > \theta\right\}
  \label{eq:hit}
\end{equation}

\normalsize
\subsection{Retriever Metrics}

\paragraph{Recall@$K$.}
Fraction of sub-queries for which at least one top-$K$ chunk is relevant, where relevance is defined as either exceeding a cosine similarity threshold of 0.35 with the gold-answer embedding or containing the gold answer as a substring:

\scriptsize
\begin{multline}
  \text{Recall@}K = \frac{1}{|Q|} \sum_{q \in Q} \mathbb{1}\bigg\{\max_{c \in \text{top}_K(q)} \cos\!\bigl(\phi(q + \text{gold}),\, \\ \phi(c)\bigr) > 0.35
  \vee\; \text{gold} \subseteq c\bigg\}
  \label{eq:recall}
\end{multline}

\normalsize
\paragraph{Precision@$K$.}
Fraction of all retrieved chunks (across all sub-queries) that are relevant to the gold answer, using a stricter cosine threshold of 0.5:

\scriptsize
\begin{equation}
  \text{Precision@}K = \frac{|\{(q, c) : c \in \text{top}_K(q),\; \cos(\phi(\text{gold}), \phi(c)) > 0.5\}|}{|Q| \times K}
  \label{eq:precision}
\end{equation}

\normalsize
\paragraph{MRR.}
Mean Reciprocal Rank over sub-queries. Let $r(q)$ be the rank of the first chunk whose cosine similarity with the gold-answer embedding exceeds 0.5 ($r(q) = \infty$ if no chunk qualifies):
\begin{equation}
  \text{MRR} = \frac{1}{|Q|} \sum_{q \in Q} \frac{1}{r(q)}
  \label{eq:mrr}
\end{equation}

\paragraph{NDCG@$K$.}
Normalized Discounted Cumulative Gain over the top-$K$ chunks. A chunk at rank $i$ is relevant ($\text{rel}_i = 1$) if its cosine similarity to the gold-answer embedding exceeds 0.5, and $R$ denotes the total number of relevant chunks:
\begin{equation}
    \begin{aligned}
  \text{DCG@}K  &= \sum_{i=1}^{K} \frac{\text{rel}_i}{\log_2(i+1)}, \\
  \text{IDCG@}K &= \sum_{i=1}^{R} \frac{1}{\log_2(i+1)}, \\
  \text{NDCG@}K &= \frac{\text{DCG@}K}{\text{IDCG@}K}
  \label{eq:ndcg}
\end{aligned}
\end{equation}


\subsection{Solver Metrics}

\paragraph{Accuracy.}
Binary correctness of the predicted answer. The Solver outputs a letter $\hat{y} \in \{\texttt{A}, \texttt{B}, \texttt{C}, \dots\}$, which is compared against the gold index $y^*$ (zero-indexed):
\begin{equation}
  \text{Accuracy} = \mathbb{1}\!\left\{\text{ord}(\hat{y}) - \text{ord}(\texttt{A}) = y^*\right\}
  \label{eq:accuracy}
\end{equation}

\paragraph{Text Citation Precision (NLI-based).}
Measures how well each text-based citation is supported by its linked evidence. Let $\mathcal{C}_\text{text}$ be the set of (evidence chunk, cited claim) pairs extracted from the Solver's response. Each pair receives a score of 1.0 (entailment), 0.5 (neutral), or 0.0 (contradiction) from the NLI model:

\scriptsize
\begin{equation}
  \text{TextCitePrec} = \frac{1}{|\mathcal{C}_\text{text}|} \sum_{(e,\,c)\,\in\,\mathcal{C}_\text{text}} \text{score}_{\text{NLI}}(e, c)
  \label{eq:text_cite_prec}
\end{equation}

\normalsize
\paragraph{Question Image Citation Precision.}
Fraction of image citations in the Solver's response that reference a valid image index. Let $\mathcal{C}_\text{QI}$ be the set of image citations and $|\text{imgs}|$ the number of images associated with the question:

\scriptsize
\begin{equation}
  \text{QICitePrec} = \frac{|\{c \in \mathcal{C}_\text{QI} : 1 \leq c.\text{id} \leq |\text{imgs}|\}|}{|\mathcal{C}_\text{QI}|}
  \label{eq:qi_prec}
\end{equation}

\normalsize
\paragraph{Citation Recall.}
Measures whether factual sentences in the Solver's chain-of-thought are backed by at least one citation. Let $\mathcal{C}_\text{all}$ denote all citations and $|\text{factual\_sentences}|$ the number of factual sentences identified in the response:

\scriptsize
\begin{equation}
  \text{CiteRecall} = \min\!\left(1.0,\;\frac{|\mathcal{C}_\text{all}|}{\max(1,\,|\text{factual\_sentences}|)}\right)
  \label{eq:cite_recall}
\end{equation}

\normalsize
\paragraph{AIS (Attributable to Identified Sources).}
Fraction of factual sentences $s \in S$ in the Solver's response for which at least one evidence chunk in $\mathcal{E}$ provides entailment support above a threshold of 0.35:

\scriptsize
\begin{equation}
  \text{AIS} = \frac{1}{|S|} \sum_{s \in S} \mathbb{1}\!\left\{\max_{e \in \mathcal{E}} P_{\text{NLI}}(\text{entailment} \mid e,\,s) > 0.35\right\}
  \label{eq:ais}
\end{equation}

\normalsize
\paragraph{Evidence Grounding Score.}
Measures how strongly the final answer $a$ is entailed by the cited evidence. It takes the maximum entailment probability over all cited chunks:
\begin{equation}
  \text{Grounding} = \max_{e \in \mathcal{E}_\text{cited}} P_{\text{NLI}}(\text{entailment} \mid e,\, a)
  \label{eq:grounding}
\end{equation}

\paragraph{CaVeScore (Composite).}
A weighted composite metric that balances answer correctness with citation and grounding quality:
\begin{multline}
  \text{CaVeScore} = w_1 \cdot \text{Accuracy} + w_2 \cdot \text{CitePrecision} \\ + w_3 \cdot \text{CiteRecall} + w_4 \cdot \text{AIS} \\ + w_5 \cdot \text{Grounding}
  \label{eq:cave}
\end{multline}
\noindent The weights $w_1, \ldots, w_5$ satisfy $w_i \geq 0$,
$\sum_i w_i = 1$, and are set to $(0.4, 0.2, 0.2, 0.1, 0.1)$ in all
experiments; see Section~\ref{app:sensitivity} for a sensitivity analysis.
The weights reflect a design priority. Correctness is necessary but insufficient; a response must also be verifiably grounded.

\subsection{Verifier Quality Metrics}

\paragraph{Ground Truth Construction.}
To evaluate the Verifier, we construct a pseudo ground-truth acceptance label from upstream metrics. A response is deemed acceptable if it is correct, well-cited, and attributable:

\small
\begin{multline}
  \text{gt\_accept} = (\text{Accuracy} = 1.0) \wedge (\text{CitePrecision} \geq 0.7) \\ \wedge (\text{AIS} \geq 0.5)
  \label{eq:gt_accept}
\end{multline}

\normalsize
\paragraph{Decision Correctness.}
Binary agreement between the Verifier's \textsc{Verified}/\textsc{Rejected} verdict and the pseudo ground-truth label $\text{gt\_accept}$.

\paragraph{Hallucination Detection Correctness.}
Binary agreement between the Verifier's hallucination flag and the NLI-derived hallucination signal, which fires when the hallucination rate exceeds 0.3.

\paragraph{Confidence Appropriateness.}
Evaluates whether the Verifier's confidence level (\textsc{High}, \textsc{Medium}, \textsc{Low}) is consistent with the evidence quality. \textsc{High} is appropriate only when citations are valid \textit{and} reasoning is grounded; \textsc{Low} is appropriate only when either condition fails; \textsc{Medium} is always considered appropriate.

\paragraph{Feedback Quality.}
Measures whether the Verifier's structured feedback correctly identifies real issues and provides actionable guidance. For each of three issue dimensions $d$ (accuracy, citation quality, grounding), let $\delta_d \in \{-1, 0, 1\}$ score true positives ($+1$), true negatives ($0$), and false positives/negatives ($-1$):

\scriptsize
\begin{equation}
\begin{aligned}
      \text{IssueAcc} &= \frac{\frac{1}{3}\sum_d \delta_d + 1}{2} \in [0,1], \\
  \text{FeedbackScore} &= 0.7 \cdot \text{IssueAcc} + 0.3 \cdot \mathbb{1}\{\text{actionable}\}
  \label{eq:feedback}
\end{aligned}
\end{equation}

\normalsize
Feedback quality is capped at 0.5 when no real issues exist, penalising spurious rejections.

\subsubsection{CaVeScore Weight Sensitivity}
\label{app:sensitivity}

The CaVeScore weights in Equation~\ref{eq:cavescore_main} encode a specific design priority. To verify that our ablation rankings are not artifacts of this particular weighting, we re-score all per-sample metric components under seven alternative weight configurations: the published default (0.4/0.2/0.2/0.1/0.1), accuracy-heavy (0.6/0.1/0.1/0.1/0.1), citation-heavy (0.2/0.3/0.3/0.1/0.1), uniform (0.2/0.2/0.2/0.2/0.2), AIS-heavy (0.3/0.15/0.15/0.25/0.15), grounding-heavy (0.3/0.15/0.15/0.15/0.25), and recall-skewed (0.35/0.1/0.35/0.1/0.1). No re-inference is required; each configuration re-weights the five pre-computed component scores and produces a new composite CaVeScore per sample.

\begin{table*}[t]
\centering
\renewcommand{\arraystretch}{1.2}
\setlength{\tabcolsep}{6pt}
\tiny
\begin{tabular}{lccccccccc}
\toprule
\textbf{Config} & $w_\text{acc}$ & $w_\text{cprec}$ & $w_\text{crec}$ & $w_\text{ais}$ & $w_\text{gnd}$ & \textbf{Solver-Only} & \textbf{Retriever-Solver} & \textbf{CaVe-VLM-CoT} & \textbf{Gap}\,$\uparrow$ \\
\midrule
default          & 0.40 & 0.20 & 0.20 & 0.10 & 0.10 & 0.314 & 0.404 & \tblbest{0.566} & $+0.251$ \\
accuracy-heavy   & 0.60 & 0.10 & 0.10 & 0.10 & 0.10 & 0.460 & 0.508 & \tblbest{0.661} & $+0.201$ \\
AIS-heavy        & 0.30 & 0.15 & 0.15 & 0.25 & 0.15 & 0.265 & 0.371 & \tblbest{0.510} & $+0.245$ \\
recall-skewed    & 0.35 & 0.10 & 0.35 & 0.10 & 0.10 & 0.278 & 0.362 & \tblbest{0.504} & $+0.226$ \\
grounding-heavy  & 0.30 & 0.15 & 0.15 & 0.15 & 0.25 & 0.263 & 0.356 & \tblbest{0.489} & $+0.227$ \\
uniform          & 0.20 & 0.20 & 0.20 & 0.20 & 0.20 & 0.191 & 0.311 & \tblbest{0.452} & $+0.261$ \\
citation-heavy   & 0.20 & 0.30 & 0.30 & 0.10 & 0.10 & 0.168 & 0.300 & \tblbest{0.470} & $+0.302$ \\
\bottomrule
\end{tabular}
\par\vspace{2pt}
{\footnotesize Gap\,$\uparrow$ = \sys{CaVe-VLM-CoT} $-$ \sys{Solver-Only}.}
\caption{CaVeScore weight sensitivity analysis across the three ablation configurations on ScienceQA ($n{=}1{,}000$). Each row re-scores every per-sample component under a different weight vector. The absolute values shift, but the ordering $\sys{Solver\text{-}Only} < \sys{Retriever\text{-}Solver} < \sys{CaVe\text{-}VLM\text{-}CoT}$ is preserved under every weighting, confirming that the ablation rankings are robust to reasonable weight perturbation.}
\label{tab:sensitivity}
\end{table*}

\noindent Two observations are worth highlighting. First, the absolute CaVeScore shifts as expected when weights change: configurations that up-weight accuracy (where our system is strongest, $87.1\%$) score higher in the aggregate, while those that up-weight the grounding-related metrics score lower. This movement is not a weakness of the metric, it is the diagnostic signal we designed CaVeScore to produce. Second, and more importantly, the \emph{relative} ordering of the three configurations is preserved under every weighting, and the gap between the full pipeline and the \sys{Solver-Only} baseline ranges from $+0.201$ (accuracy-heavy) to $+0.302$ (citation-heavy). The gap widens precisely under the weightings that stress grounding quality, which is the regime where the full pipeline's retrieval and verification machinery matters most. The ablation conclusions therefore do not depend on the specific choice of weights in Equation~\ref{eq:cavescore_main}.

\section{Experimental Results}
\label{sec:full_results}

Table~\ref{tab:full_results} reports all 23 metrics as mean $\pm$ standard deviation over
$1{,}000$ ScienceQA examples.
 
\section{Implementation Details}
\label{sec:implementation}

\paragraph{Dataset preprocessing.}
All ScienceQA images are resized to $224 \times 224$ pixels and converted to RGB.
Captions are generated using BLIP~\citep{li2022blip}
(\texttt{Salesforce/blip-image-captioning-base}) in batches of eight. OCR is performed
using Tesseract. The processed dataset is serialised as both CSV and JSON, with image
paths updated to resized copies, so every pipeline stage reads from the same augmented
representation.

\paragraph{MMMU dataset details.}
MMMU~\citep{yue2024mmmu} covers 30 subjects (e.g., medicine, engineering, art 
history, economics) with $500$ questions across \texttt{dev} and   
\texttt{validation} splits. It differs from ScienceQA in three respects: 
(i)~every question includes at least one figure (chart, diagram, microscopy 
image, or blueprint), making visual reasoning mandatory; (ii)~questions require 
university-level domain expertise, testing whether the retrieval pipeline can 
surface specialised evidence; and (iii)~no built-in lecture corpus exists, so 
the Retriever relies more heavily on web search. Images are downloaded from 
HuggingFace Hub via the \texttt{datasets} library and resized to 
$224 \times 224$ pixels. BLIP captions are generated identically to ScienceQA; 
OCR is omitted as MMMU images are predominantly diagrams and charts. The 
\texttt{explanation} field (present only in the \texttt{dev} split) is mapped 
to the \texttt{lecture} column for knowledge base indexing; \texttt{validation} 
split rows receive an empty \texttt{lecture} field. MMMU's FAISS index is 
stored in a separate subdirectory (\texttt{indexes/mmmu/}) to avoid overwriting 
the ScienceQA index. Answer letters (\texttt{A}-\texttt{E}) are converted to 
zero-indexed integers. No pipeline code changes were required.

\paragraph{Knowledge base indexing.}
Dense embeddings use \texttt{all-MiniLM-L6-v2}~\citep{reimers2019sentence} (384
dimensions), normalised to unit length and stored in a FAISS flat
index~\citep{johnson2019billion} for exact inner-product search. The sparse index uses
BM25Okapi with whitespace tokenisation. Hybrid fusion uses RRF with $k=60$.

\paragraph{Retrieval hyperparameters.}
Dense and sparse retrieval each return top-$3$ candidates per sub-query before fusion.
The cross-encoder reranker (\texttt{cross-encoder/ms-marco-MiniLM-L-6-v2}) operates over
a top-50 candidate pool; the top two local corpus chunks are reserved unconditionally and
up to $k{=}5$ total chunks are passed per sub-query (10 maximum across all sub-queries).
Web search returns up to two snippets per sub-query plus one additional snippet per
choice-augmented variant.

\paragraph{Model configuration.}
The Extractor (Qwen2.5-7B-Instruct) is loaded in 4-bit NF4 quantisation via \texttt{bitsandbytes}
(\texttt{unsloth/Qwen2.5-7B-Instruct-bnb-4bit}), with greedy decoding ($T{=}0.0$) and a maximum
of 512 new tokens. The Solver (LLaVA-CoT~\citep{xu2025llava}, an 11B-parameter Llama-3.2-Vision model fine-tuned for chain-of-thought reasoning; checkpoint \texttt{zhangsongbo365/Llama-3.2V-11B-cot-nf4})
uses greedy decoding ($T{=}0.0$) and a maximum of 1024 new tokens; citation-retry calls
use $T{=}0.1$. The Verifier (Qwen2.5-VL-32B-Instruct) uses greedy decoding ($T{=}0.0$)
and a maximum of 2048 new tokens, ensuring deterministic verdicts. All three models run
on dedicated NVIDIA A40 48\,GB GPUs (one model per GPU) with PyTorch's math SDPA
attention backend.

\paragraph{NLI evaluation.}
All NLI-based metrics use \texttt{cross-encoder/nli-deberta-v3-base} with
\texttt{apply\_softmax=True}. Entailment threshold for AIS is 0.35; 
and
precision metrics, 0.5.

\paragraph{Observability.}
Every pipeline stage is instrumented via OpenInference-compliant spans capturing model
inputs, outputs, token counts, latencies, and per-stage evaluation metrics, enabling
post-hoc per-example trace analysis.

\section{Comparison with Existing Evaluation Frameworks}
\label{app:metric_comparison}

Table~\ref{tab:metric_comparison} contrasts the quality dimensions 
captured by CaVeScore against those of existing evaluation approaches.
We mark a dimension as \checkmark\ when it is directly measured by the 
framework, $\sim$ when partially captured, and $\varnothing$ when not 
measured. Judgements are made on each framework's primary or headline 
metric; frameworks often report auxiliary numbers that touch adjacent 
dimensions, but these are not the axes on which they are designed to 
rank systems.

\begin{table*}[t]
\centering
\renewcommand{\arraystretch}{1}
\setlength{\tabcolsep}{5pt}
\small
\begin{tabular}{lccccc}
\toprule
\textbf{Metric / Framework} & \makecell{\textbf{Answer}\\\textbf{Correct.}} & \makecell{\textbf{Citation}\\\textbf{Precision}} & \makecell{\textbf{Citation}\\\textbf{Recall}} & \makecell{\textbf{Step-level}\\\textbf{Attribution}} & \makecell{\textbf{Evidence}\\\textbf{Grounding}} \\
\midrule
Accuracy / Exact Match & \checkmark & $\varnothing$ & $\varnothing$ & $\varnothing$ & $\varnothing$ \\
BLEU / ROUGE & $\sim$ & $\varnothing$ & $\varnothing$ & $\varnothing$ & $\varnothing$ \\
ALCE~\citep{gao2023alce} & $\varnothing$ & \checkmark & \checkmark & $\varnothing$ & $\varnothing$ \\
AIS~\citep{ji2024chain} & $\varnothing$ & $\varnothing$ & $\varnothing$ & \checkmark & $\varnothing$ \\
LLaVA-Critic~\citep{xiong2025llavacritic} & $\sim$ & $\varnothing$ & $\varnothing$ & $\varnothing$ & $\sim$ \\
VHELM~\citep{lee2024vhelm} & \checkmark & $\varnothing$ & $\varnothing$ & $\varnothing$ & $\varnothing$ \\
\rowcolor{bestrow}
\textbf{CaVeScore (ours)} & \checkmark & \checkmark & \checkmark & \checkmark & \checkmark \\
\bottomrule
\end{tabular}
\caption{Comparison of evaluation metrics across quality dimensions. \checkmark = directly measured; $\sim$ = partially captured; $\varnothing$ = not measured.}
\label{tab:metric_comparison}
\end{table*}

\end{document}